\documentclass{article}
\PassOptionsToPackage{numbers, square}{natbib}

    \usepackage[final]{neurips_2024}

\usepackage[utf8]{inputenc} 
\usepackage[T1]{fontenc}    
\usepackage{hyperref}       
\usepackage{url}            
\usepackage{booktabs}       
\usepackage{amsfonts}       
\usepackage{nicefrac}       
\usepackage{microtype}      
\usepackage{xcolor}         
\usepackage{graphicx}

\title{Diverse capability and scaling of diffusion and auto-regressive models when learning abstract rules}
\author{Binxu Wang\\ 
Kempner Institute, Harvard University\\ 
\texttt{binxu\_wang@hms.harvard.edu}\\
  \And {Jiaqi Shang}\\ 
  Program in Neuroscience, Harvard Medical School\\
  \texttt{jiaqishang@g.harvard.edu}\\
\And {Haim Sompolinsky} \\
  Center for Brain Science, Harvard University\\
Edmond and Lily Safra Center for Brain Sciences, Hebrew University\\
\texttt{hsompolinsky@mcb.harvard.edu}}

\begin{document}

\maketitle

\begin{abstract}
  Humans excel at discovering regular structures from limited samples and applying inferred rules to novel settings. We investigate whether modern generative AI systems can similarly learn underlying rules from finite samples and perform reasoning through conditional sampling. 
  Inspired by Raven’s Progressive Matrices task, we designed GenRAVEN dataset, where each sample consists of three rows, and one of the 40 relational rules governing the object position, number, or attributes applies to all three rows. 
  We trained generative models to learn the data distribution, where samples are encoded as 3×9×9 integer arrays to focus on rule learning. 
  We compared two major families of generative models: diffusion models (EDM, DiT, SiT) and autoregressive models (GPT2, Mamba). We evaluated their ability to generate structurally consistent samples and perform panel completion via unconditional and conditional sampling. 
  We found that diffusion models excel at unconditional generation, producing novel and more consistent samples from scratch and memorize less, but perform less well in panel completion, even with advanced conditional sampling methods like Twisted Diffusion Sampler. Conversely, autoregressive models excel at completing missing panels in a rule-consistent manner but generate less consistent samples unconditionally. 
  We observe diverse data scaling behaviors: for both model families, rule learning emerges at a certain dataset size -- around thousands examples per rule. With more training data, diffusion models improve both their unconditional and conditional generation capabilities. However, for autoregressive models, while panel completion improves with more training data, unconditional generation consistency declines. 
  Our findings highlight complementary capabilities and limitations of diffusion and autoregressive models in rule learning and reasoning tasks, suggesting avenues for further research into their mechanisms and potential for human-like reasoning.
\end{abstract}

\section{Background}

\begin{figure}[!bth]
  \centering
  \includegraphics[width=\textwidth]{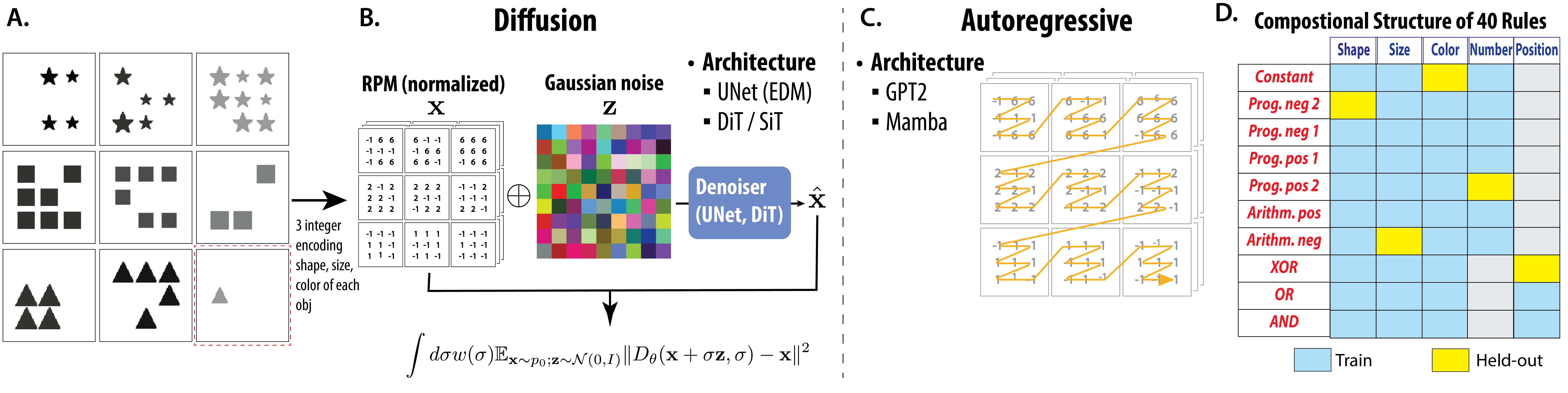}
  \caption{\textbf{Design of the study} \textbf{A.} Example Raven's progression matrix, and its encoding as a 3$\times$9$\times$9 integer array. The underlying rule is \textcolor{red}{constant} \textcolor{blue}{shape}. \textbf{B.C.} Two families of generative models: Diffusion and autoregressive model, and their training method: denoising and predicting the next token. \textbf{D.} The 40 relational rules, with 5 rules held out during training.}\label{fig:method_schematics}
\end{figure}

Human excels at discovering regular structure from a small number of samples, and they can further apply such rule to novel settings to generate new samples or complete missing parts based on the same rule. The Raven's progressive matrix (RPM) \cite{raven1936mental} is a famous task in human reasoning literature. In the generative version of this task (GenRAVEN), the subject observes two complete rows of panels and is tasked to complete the third row in a manner that is consistent with the first two rows (Fig.\ref{fig:method_schematics}A). Ideally, the subjects need to infer the underlying rule consistent with the first two rows and apply it correctly to the third row. 
How can we train a general learning system to solve such reasoning task?  

If we conceptualize all rule-conforming samples as a joint distribution, then rule learning can be framed as a generative modeling problem or learning the correct joint. Further, reasoning about the missing panel can be framed as sampling from the conditional probability \cite{prystawski2024whycot}. 
One conceptual problem is, that given finite training samples, the rule governing them, or the `true` joint distribution is under-specified. The rules or the distribution learned by the system should be affected by its inductive bias, be it human or AI. 
Given this ambiguity, we asked whether modern generative AI systems could learn the correct "joint distribution" given finite samples. If so, can they reason and fill in the missing parts in a sample through conditional sampling? 

In the current age of Generative AI, there are two prominent families of generative models: autoregressive models and diffusion models. The autoregressive model generally dominates discrete sequence data, such as language, music, genomics \cite{brown2020GPT3,hyenadna2023genomic,huang2018musictransformer}, while the diffusion model excels at continuous data, such as image, video, audio, molecule structure, robot trajectories \cite{rombach2022StablDiff,wu2022diffMolecule,ruan2023mmDiff,carvalho2023motionDiff,blattmann2023alignlatentvideo}. Both of them are capable of unconditional generation and conditional sampling based on partial observation: e.g. prompting for autoregressive model or inpainting for diffusion models. Given their similar capability, it's interesting to compare them back-to-back on the same reasoning task and see how their different modeling method might lead to different learning and scaling behaviors.  


In this work, we trained a diverse set of generative models from both the diffusion family (EDM, DiT, SiT) and autoregressive family (GPT, Mamba) to learn the data distribution and then use conditional sampling techniques to perform inference, i.e., reasoning about the missing panel. We studied their performance as a function of data scale and model scale. We found that generally, diffusion models excel at unconditional sampling from the joint, creating structurally consistent samples from scratch, but perform less well for sampling conditioned on given panels. Conversely, autoregressive models excel at completing missing panels in a rule-consistent manner, but they perform less well in generating consistent samples from scratch with unconditional sampling. They also exhibit diverse scaling behavior when the size of datasets is varied. Our results call for further investigation into the seemingly complementary capabilities and limitations of diffusion and autoregressive models. 

\section{Method}
\paragraph{GenRAVEN Dataset}
We introduce the GenRAVEN dataset, comprising RPMs associated with 40 relational rules. Each RPM features a 3$\times$3 matrix of panels, where the three panels in each row follow a unique relational rule. 
To focus on rule learning and reasoning, we abstracted away the visual aspect of the sample, encoding each object with three integers representing its shape, size, and color, with $(-1,-1,-1)$ representing empty positions. Each attribute has a discrete set of allowed values, 0-6 for shape and 0-9 for size and color. Taking together, each sample RPM is represented by an integer array of shape $3\times9\times9$, which is the target of generative modeling (Fig. \ref{fig:method_schematics}B). 

Each rule is composed of an \textcolor{red}{abstract relation} (\textcolor{red}{constant; progression ±2, ±1; arithmetic ±; XOR; OR; AND}) applied to an \textcolor{blue}{attribute} (\textcolor{blue}{shape, size, color, number, position}). 
The dataset is designed so that the rule governing each row remains ambiguous when examining only the first two panels and only becomes evident when all three panels are considered. This design ensures that the rule governing the row cannot be directly deduced from the first two panels alone, and the model must reason the entire matrix. 
We generate 4000 random samples per rule for training. 
To study the generalization of abstract relations such as \textcolor{red}{constant} to new attributes, we held out 5 rules (Fig \ref{fig:method_schematics}D) during training. 

\paragraph{Diffusion Model} 
The diffusion models have been a prominent approach for generative modeling \cite{dhariwal2021diffbeatsGAN} in continuous domains. 
It takes a holistic approach to modeling: it learns a vector field in the data space $s_\theta(x,\sigma)$, approximating the gradient of smoothed data distribution. When states flow along the vector field, they will be transported from a base distribution, e.g., Gaussian, into the target distribution $p(X)$. During sampling, samples initialized from Gaussian will be transformed via this dynamic process into generated samples. 

Given the spatial structure of the task, we treated each RPM as a 9$\times$9 image with 3 attribute channels and adapted existing diffusion models for image generation. 
Specifically, we experimented with two network architectures, UNet \cite{karras2022elucidatingEDM} and Diffusion Transformer (DiT) \cite{peebles2023scalableDiT}. 
UNet is a Convolutional Neural Network (CNN) based backbone designed to extract information at multiple resolutions, with self-attention modules at each resolution. For a long time, it has been the default backbone for diffusion models for images \cite{ho2020DDPM,dhariwal2021DIFbeatGAN,rombach2022StablDiff}. For our task, we adapted its architecture to match the 3x3 panel structure of the RPM samples: we changed the filter size to 3 and the downsampling ratio to 3. 
DiT is a recent transformer-based backbone adapted for diffusion modeling \cite{peebles2023scalableDiT}. For our task, we treated each object in RPM as a token with three features (patch size = 1), totaling 81 tokens. 
We also tested the SiT model \cite{ma2024sit}, which had the same network architecture as DiT, but based their training and inference on the theoretical framework of Stochastic Interpolant \cite{albergo2023stochasticinterpolants}. 


We treated the integer attributes in RPM samples as continuous values and normalized them by mean and std before training. During sampling, since diffusion models generate samples with continuous values, we rounded the generated attribute value to the closest integer. 
We used deterministic samplers to generate samples: Heun's 2nd order sampler with 18 steps for UNet model \cite{karras2022elucidatingEDM}; DDIM sampler with 100 steps for DiT \cite{song2020DDIM}; off-the-shelf Runge-Kutta sampler of order 5 (\texttt{dpori5}) for SiT \cite{ma2024sit}. 
Leveraging the in-painting capability of diffusion models \cite{lugmayr2022repaint,wang2022DDNM}, we also challenged them with the RPM inference tasks: given the first 8 panels, let the model sample the missing panel. 

\paragraph{Autoregressive models} 
Another major family of generative models is the autoregressive models, which have demonstrated impressive capabilities to model complex sequence data like natural language and music \cite{brown2020GPT3,huang2018musictransformer}. 
These models have also been applied to model natural images as sequences of patches and quantized tokens \cite{esser2021tamingVQGAN,chang2022maskgit}. These models decompose the joint distribution into the conditional probability of each variable over all previous variables $p(X)=p(x_1)p(x_2|x_1)p(x_3|x_{1:2})...p(x_d|x_{1:d-1})$. During inference, it samples variables autoregressively as a sequence. 
Here we used auto-regressive models \textbf{GPT2} and \textbf{Mamba} \cite{radford2019language,gu2024mamba} to solve RAVEN's task, learning the sequence of objects in an RPM. 
Similar to DiT, we treated each object as a token. The three integer attributes of each token were embedded through separated embedding layers with 1/3 hidden dimensions and then concatenated as the token embedding. To reflect such latent space structure, the hidden state outputs from the transformer or Mamba were chunked into 3 parts and decoded separately into attributes of the next token. We used the default learned positional encoding for GPT2 and no position encoding for Mamba.  
We train these models with the next token prediction objective. A key design choice is the order to re-arrange 2 dimensional RPM data into a one-dimensional sequence. Here we first scan the objects within each panel and then follow the raster order of panels (panel 1 row 1, panel 2 row 1, ...panel 3 row 3), forming a sequence of 81 tokens (see Fig.\ref{fig:method_schematics}C). We prepend the same starting token \texttt{[SOS]} with learned embedding for each sample. 
For unconditional generation, we started from \texttt{[SOS]} and autoregressively sampled 81 tokens with a temperature of 1.0. For panel completion, we started from \texttt{[SOS]} and 72 existing tokens (corresponding to the first 8 panels), and sampled 9 tokens with temperature 1.0. 

\paragraph{Evaluation} 
One benefit of having a distribution defined by explicit rule is that we can evaluate the generated samples more precisely. 
For unconditional generation, there is no sense of accuracy, so we evaluated the internal consistency of generated samples. For each row in a generated sample, we inferred the set of applicable rules. If any rule from the 40 rule set applies, the row is called \textit{valid}. We calculated the fraction of valid rows throughout training. For each sample, we examined whether the same rule applies to two or three rows in the sample, which we call the two or three-row consistent fraction (\textit{C2, C3}). 

For conditional generation, we selected 50 new samples per rule, unseen during training, as our test cases. We removed their final panel and let the model generate the missing panel given the other 8 panels. We evaluated whether the completed panel allowed the same rule to apply to all three rows (C3). The accuracy of panel completion was quantified via the frequency of C3 completion.   

\section{Results}

\subsection{Diffusion models learn to generate structurally consistent samples better}
\begin{figure}[!bth]
  \centering
  \vspace{-10pt}
  \includegraphics[width=0.99\textwidth]{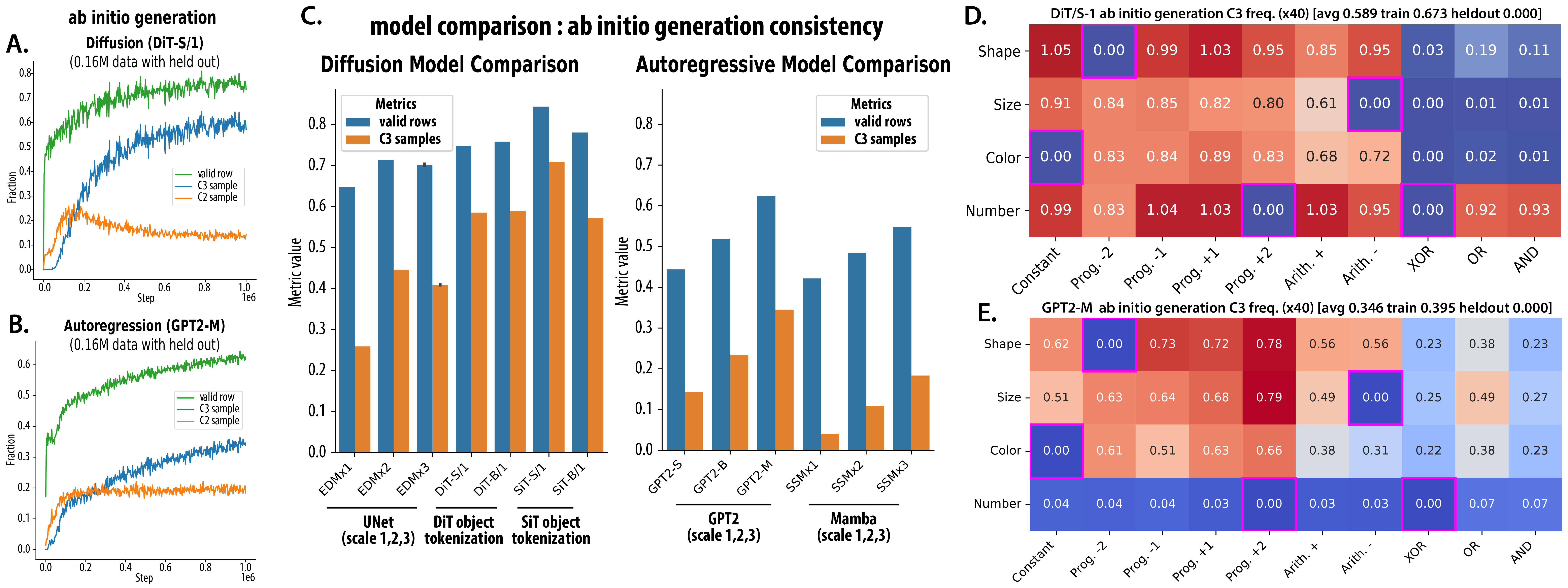}
  \vspace{-7pt}
  \caption{\textbf{Diffusion models lead in generating structurally consistent samples \textit{ab initio}} 
  \textbf{A.B.} Dynamics of sample consistency per valid row fraction and C2,C3 fraction during training, for diffusion model (\textbf{A.} DiT-S/1) and autoregressive model (\textbf{B.} GPT2) 
  \textbf{C.} Comparison of \textit{ab initio} generation consistency across model families. 
  \textbf{D.E.} Frequency of generating C3 samples of each rule (showing the value $\times$40 to normalize) for DiT-S and GPT2-M. Magenta frames showing the 5 rules held out from generative model training.  }\label{fig:uncond_generation}
  \vspace{-5pt}
\end{figure}

\paragraph{Diffusion models excel in \textit{ab initio} generation }
We found that, through generative modeling, both diffusion and autoregressive models learn to generate structurally consistent samples measured via valid row fraction and C3 sample fraction (Fig. \ref{fig:uncond_generation}A,B). 
Comparing across model families (Fig. \ref{fig:uncond_generation}C), SiT models perform better than DiT, and EDM models, which generally perform better than GPT and Mamba models. Notably, the best-performing diffusion model SiT-S/1 had 70.9\% of generated samples that were C3 rule consistent; in contrast, the best-performing auto-regressive model (GPT2-M) only had 34.6\% C3 samples while doubling the layer number and hidden space dimension of SiT-S. 
Within each model family, we found scaling up model size helps only for lower performing families, e.g., EDM, GPT, and Mamba, but for DiT and SiT models, it didn't help or even hurt performance. The limited effect of model scaling may be related to a fixed training set size, which we will manipulate below (Fig.\ref{fig:diverse_scaling}). 
Notably, across model families, SiT-S/B, DiT-S/B, and GPT2-S/B form nice comparison pairs, since they match in their number of attention heads, layers, hidden dimension, tokenization, and training steps. Given this fair comparison, DiT models have a higher rate of generating structurally consistent samples than GPT2. 

\paragraph{Different rule types were challenging for each family of models} 
When separated by the rule types, the two classes of models fell short at different rule types (Fig.\ref{fig:uncond_generation}\textbf{D, E}). 
Diffusion models rarely generate consistent samples with logical operation applying to attribute sets (e.g. Shape-XOR, which means when the first panel contains circles and triangles, the second panel contains triangles and squares, then the third panel could only contain circles and squares.) For the held-out rules, the models did not generate C3 samples consistent with those rules after training, though single rows consistent with held-out rules were still generated with some frequency. 
For Autoregressive models, they failed spectacularly for rules related to number and position (e.g. Prog.+1-Number, which means 3 objects in the first panel, 4 objects in the 2nd panel, 5 in the 3rd panel; Position-AND, which means 3rd panel can only place objects at locations where objects exist in both 1st and 2nd panel). They had a higher frequency of generating according to logic-attribute rules, compared to diffusion models. Comparatively, their success rate in generating progression and arithmetic rules on attribute values is also lower. 


\subsection{Diffusion models and autoregressive model display diverse memorization behavior} 
When models create structurally consistent samples or valid rows, are they creating something completely novel or simply recalling their training set? To answer this, we evaluated the memorization properties of generated samples and their parts: rows, panels, and rows and panels when considering single attributes. We evaluated what fraction of samples and parts have exact copies in the training set for the model. As a control, we also generated another control dataset with 4000 samples per rule unseen during training and computed the memorization fraction towards the control set. 

\begin{figure}[!bth]
  \centering
  \vspace{-10pt}
  \includegraphics[width=\textwidth]{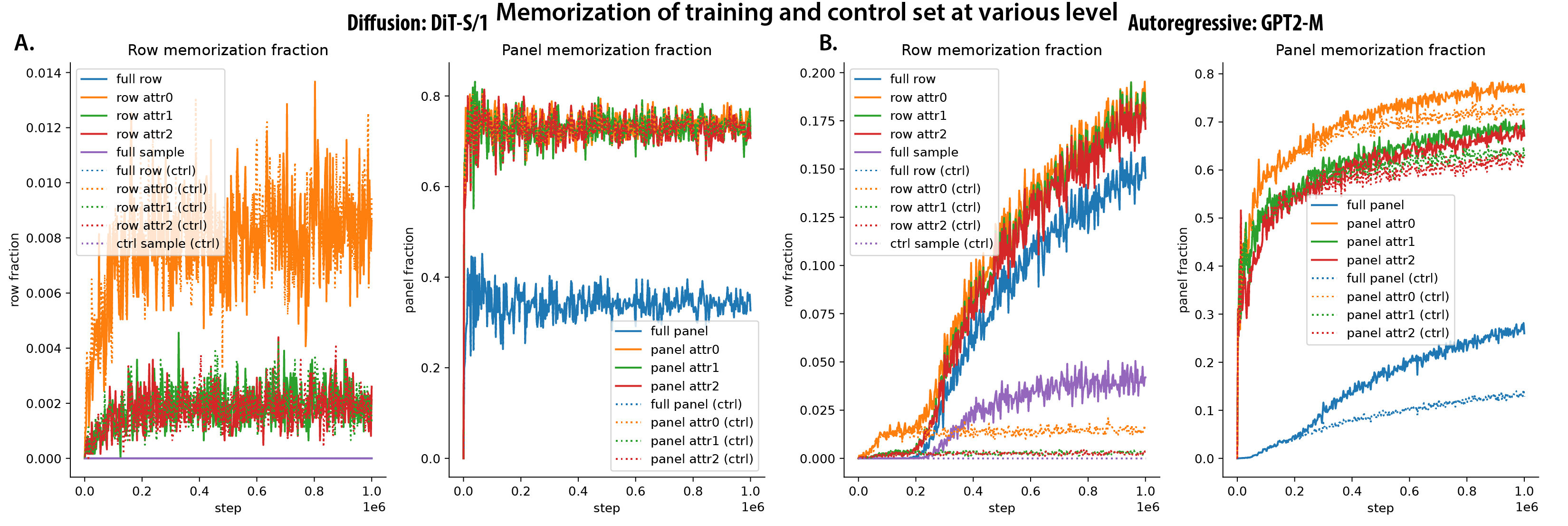}
  \vspace{-12pt}
  \caption{\textbf{Diffusion and autoregressive models show diverse data memorization property}
  \textbf{A. B.} Memorization of training and control set at multiple levels for samples generated through training, for \textbf{A.} DiT-S, and \textbf{B.} GPT2-M. Solid lines show what fraction of samples, rows, and panels have copies from the training set, and dashed lines show the control, i.e. the fraction of those with copies from the control set of samples unseen during training.  
  }\label{fig:memorization}
  \vspace{-5pt}
\end{figure}

\paragraph{Diffusion models show little memorization on row and sample level}
Remarkably, within the 800,000 samples generated by the DiT-S model through training, not a single sample or row has an exact copy from the training set! Even if we only consider one attribute of a row, only 0.9\%, 0.2\%, and 0.2\% of rows generated by the DiT-S model were memorized after training. 
On the local level, 34.0\% of the generated panels were memorized, and their single attributes components have over 70\% memorization rate. Intriguingly, the "memorization" on the panel level happened very rapidly during initial training and then stabilized at a fixed level (Fig.\ref{fig:memorization}A).  

We further found the memorization rates of row and panel attributes are comparable with the control memorization rate, i.e. generated samples overlapping with the control set not used for training. Thus, we can conclude this is "benign" memorization, or approaching the true data distribution. 
The negligible deviation between the solid and dashed lines (Fig.\ref{fig:memorization}A.) highlights diffusion models are learning the true data distribution without over-focusing on the training set. 
Similar results were observed in other diffusion models. 
This shows that the diffusion models are capable of generating novel rule-consistent rows and samples without memorizing training rows, demonstrating that "understanding" of the rules is on the abstract level. Further, given its memorization of panels, it seems they are capable of recombining memorized local parts (panels, attribute panels) to create novel rows and samples consistent with rules. 

\paragraph{Autoregressive models show significantly more memorization with lower consistency}
In contrast, after training, generation from the GPT2-M model had 4.1\% of samples, 14.9\% of rows, and 26.8\% of panels that were memorized from the training set. This shows that many valid generated rows from GPT2 models are simply regurgitating their training sequences. This compounds the low generation consistency and highlights the "lack of understanding" of abstract rules in autoregressive models (Fig.\ref{fig:memorization}B.). 
Intriguingly, panels are memorized less in GPT2-M than DiT-S. Further, the panel memorization rate of GPT2-M converged to a high level much slower than DiT-S. 
Compared to the control memorization rate, we can see the memorization rate with training and control set are initially aligned and both increasing before they diverge around 0.25M steps, showing the model starts to over-memorize the specific training samples. 

This analysis highlights a dichotomy between the memorization behavior in diffusion and autoregressive models: while diffusion models tend to rapidly memorize local panels and attribute combinations, autoregressive models default to memorize more global rows and samples. Diffusion models mostly showed "benign" memorization that is consistent with the statistics of the true distribution, while autoregressive models show significant over-memorization of the training set. 


\subsection{Autoregressive models excel at rule consistent panels completion}
\begin{figure}[!bth]
  \centering
  \vspace{-10pt}
  \includegraphics[width=\textwidth]{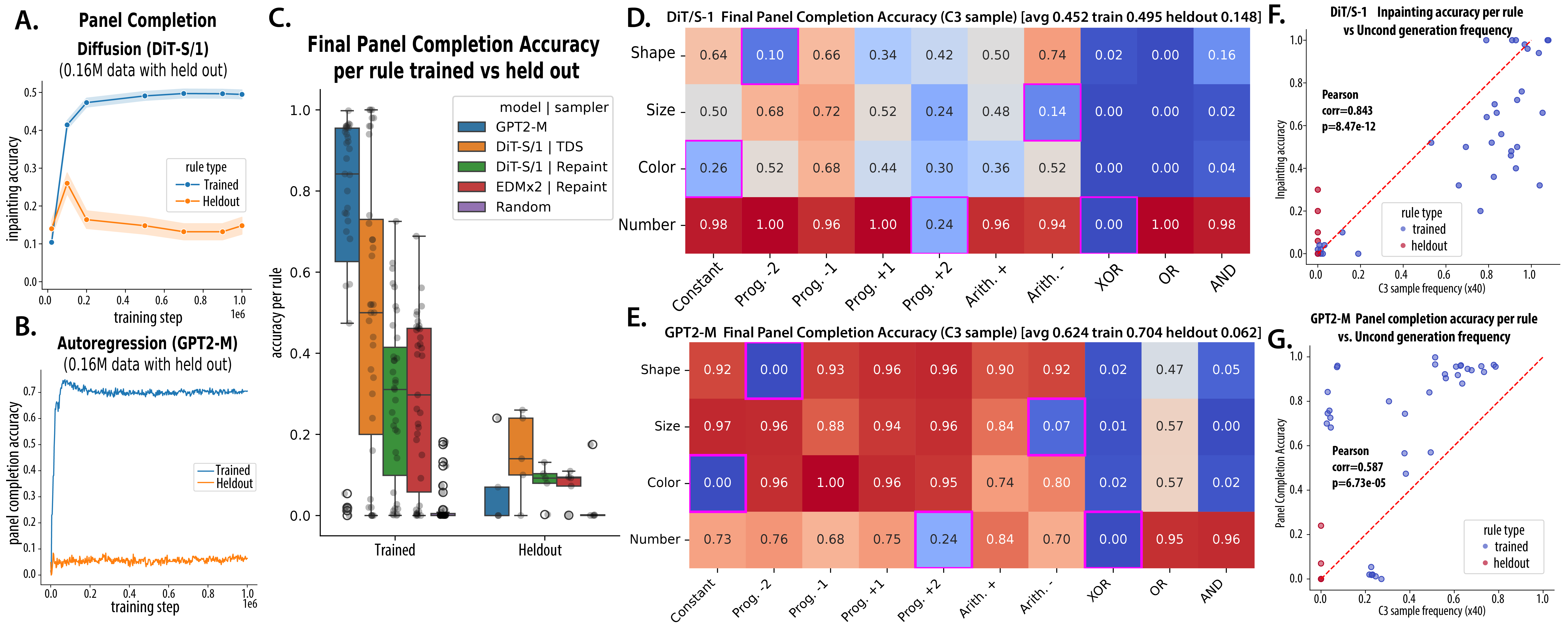}
  \vspace{-12pt}
  \caption{\textbf{Autoregressive models lead in rule consistent panel completion}
  \textbf{A. B.} Learning dynamics of panel completion accuracy for trained and held out rules, for diffusion model (\textbf{A.} DiT-S/1) and autoregressive model (\textbf{B.} GPT2) 
  \textbf{C.} Comparison of panel completion accuracy across model class and sampler. 
  \textbf{D. E.} Panel completion accuracy per rule for DiT-S and GPT2-M after training. 
  \textbf{F. G.} Correlation between panel completion accuracy and C3 sample generation frequency (Fig.\ref{fig:uncond_generation}D.E.) per rule for DiT-S and GPT2-M after training.
  }\label{fig:cond_generation}
  \vspace{-3pt}
\end{figure}

Next, we examined their capability of completing missing panels. 
Similar to unconditional generation, their panel completion accuracy also grows with training in both diffusion and autoregressive models (Fig. \ref{fig:cond_generation} \textbf{A.B.}). Notably, the learning curve of conditional generation saturates faster than unconditional generation, suggesting making a fully consistent sample from scratch (unconditional sample from joint) may be a harder task than panel completion (sample conditionally). 

\paragraph{Inpainting accuracy significantly depends on the conditional sampling method for diffusion model}
One salient observation is that, for diffusion models, their panel completion accuracy depended critically on the conditional sampling method. 
Repaint and DDNM \cite{wang2022DDNM,lugmayr2022repaint} are popular methods for general image inpainting or linear inverse problems. They are heuristic methods that basically fix the values of observed pixels and only let the masked pixels diffuse with diffusion. However, these methods do not perform exact conditional sampling \cite{albergo2023SIdata_coupling}. 
Recently, twisted diffusion sampler (TDS) \cite{wu2024TwistedDiffusionSampler} was shown to be asymptotically exact. TDS maintains a population of particles and re-weights them during sampling, leveraging the twisting technique from Sequential Monte Carlo (SMC) literature. We tested both methods on our task and found TDS performed much better than the Repaint method: on DiT-S/1 model, TDS yielded an overall C3 completion accuracy (across all 40 rules) of 45.2\% while Repaint yielded accuracy 26.1\% (Fig. \ref{fig:cond_generation} C). 

This is interesting, since for natural image completion, evaluation is usually done through visual inspection or classification, which is a relatively loose criterion. Then heuristic samplers, e.g., Repaint and DDNM, seem to suffice. But for the RPM task, where we have access to the precise form of underlying rules, the evaluation of sample completion can be done much more precisely. For such a task, advanced sampling algorithms seem to be required to perform well.  

\paragraph{Autoregressive models lead in panel completion while generalizing less well.} 
Intriguingly, even with the advanced TDS sampler, diffusion models fall short in their panel completion accuracy, falling behind GPT2 models: GPT2-M model had panel completion accuracy of 62.4\% (Fig.\ref{fig:cond_generation}C). This is the opposite of unconditional generation. Notably, for held-out rules, diffusion models with or without twisted sampler still show comparable or higher completion accuracy than chance and GPT2 models, showing that diffusion models still exhibit better rule-level generalization. 

\paragraph{Panel completion ability correlates with unconditional generation capability}
Notably, within each model, when analyzed per rules, the panel completion accuracy of each rule correlates with the frequency of C3 samples in unconditional generation  (Fig.\ref{fig:cond_generation}F.G.): Pearson correlation 0.843 ($p=8.4\times10^{-12}$) for DiT-S/1 and 0.587 ($p=6.7\times10^{-5}$) for GPT2-M. 
This suggests that a similar mechanism within the models may contribute to both panel completion and unconditional generation capabilities. 
Comparing between DiT and GPT2, note that the dots generally fall on different sides of the diagonal, where the completion accuracy per rule is generally lower than the unconditional C3 frequency for DiT, but the opposite is true for GPT2. 

\subsection{Diverse scaling property of diffusion model and autoregressive model}
Next, we studied the effect of dataset and model scales on the diffusion and autoregressive model. 
\begin{figure}[!bth]
  \centering
  \vspace{-2pt}
  \includegraphics[width=0.97\textwidth]{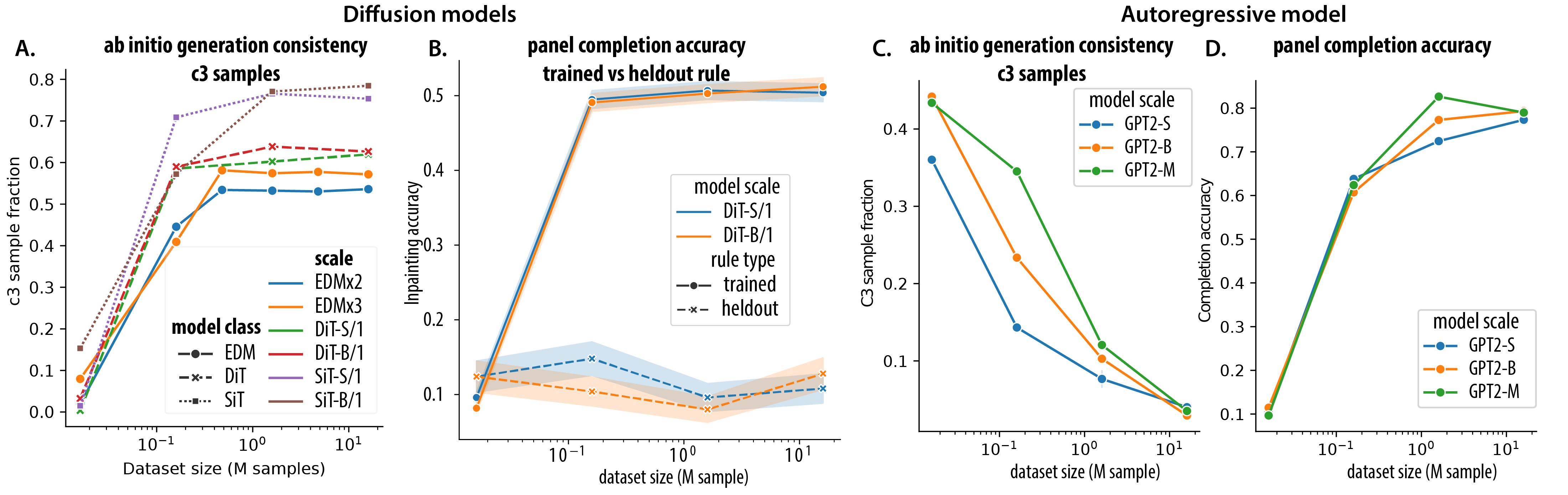}
  \vspace{-7pt}
  \caption{\textbf{Diverse scaling behavior of Diffusion and Autoregressive models}
  \textbf{A.B.} Data scaling curve of Diffusion models (EDM, DiT, SiT) \textbf{A.} \textit{ab initio} generation consistency (C3 fraction) and \textbf{B.} panel completion accuracy.
  \textbf{C.D.} Analogous data scaling curve of autoregressive model (GPT2)
  }\label{fig:diverse_scaling}
\end{figure}

\paragraph{Rule learning collapses at low data scale for diffusion model}
For diffusion models, across the three classes, we found that at a small dataset scale (400 samples per rule, or 0.016M total sample), the training "collapsed" and both the unconditional consistency and the panel completion accuracy dropped to a low level (Fig.\ref{fig:diverse_scaling}A, B). Beyond that scale, scaling up the dataset results in a higher C3 ratio and completion accuracy, with larger models benefiting more from the larger data (Fig.\ref{fig:diverse_scaling}A). We didn't see much modulation in the learning of the held out rule by dataset scale (Fig.\ref{fig:diverse_scaling}B). 
Generally, across models and across scales, the unconditional and conditional generation capabilities of diffusion models are correlated, suggesting similar mechanisms behind them. 

\paragraph{Divergent data scaling for autoregressive model} 
For autoregressive model, however, as the dataset size grows, we observed contrasting trends between the two tasks. 
Similar to diffusion models, their panel completion accuracy also collapsed at the same scale (400 samples per rule), and the accuracy grows with the data scale up to a ceiling (Fig. \ref{fig:diverse_scaling}\textbf{D}). Across model scales, larger models (GPT2-M) benefit more from larger data than smaller ones, similar to diffusion models. 
Conversely, as for unconditional generation (Fig. \ref{fig:diverse_scaling}\textbf{C}), \textit{ab initio} sample consistency decreases sharply as a function of dataset size, where smaller models (GPT2-S) decays faster. 
Thus, although the panel completion accuracy benefits from bigger training datasets, scaling up datasets has a detrimental effect on generating structurally consistent samples, esp. for smaller models. Such diverging data scaling properties of conditional and unconditional sampling capability for autoregressive models are worth further investigation. 

\section{Discussion and Future direction}

\paragraph{Weakness of autoregressive models in structurally consistent samples generation from scratch} 
Given the high panel completion accuracy of GPT models, we are surprised by the low structural consistency (valid row, C3) of their samples when we let them generate unconditionally. Why did they struggle to generate samples from scratch? 
We are reminded of an argument Yann LeCun made \cite{lecun_Talkslides_2023,lecun_twitter_2023}: even if the conditional probability of sampling each token is slightly off, autoregressive sampling will lead to exponentially compounding errors; further when one error is made during sampling, the model has no way to come back and revise it. 
In contrast, diffusion models have a holistic view of the sample from the start and can always change any token on the RPM, which may lead to higher structural consistency of the samples. It is interesting to further validate this compounding error hypothesis for the autoregressive model and see how we can rescue it with a better sampling method (e.g., beam search). 

\paragraph{Weakness of diffusion models in rule consistent panel completion} 
On the flip side, we are also surprised that the panel completion accuracy of diffusion models is generally lower than the C3 frequency of their unconditional generation. As we may naively think, modeling conditional density should be easier than modeling joint distribution. 
One potential reason is that even though unconditional sampling algorithms for diffusion have been studied extensively \cite{karras2022elucidating,zhao2024unipc,liu2022PNDM,lu2022dpm}, we still do not have the perfect method for sampling the conditional distribution from diffusion models. Just as TDS performs better than simple heuristic methods, e.g., Repaint, a better conditional sampler may boost the reasoning capability of diffusion models further. 
Our results also provide evidence that for tasks that are evaluated precisely (e.g. abstract rules), a better sampler is critical. 

\paragraph{Comparative study of rule learning with humans and animals} 
We showed that modern generative models can learn abstract rules from finite rule-conforming samples. However, their sample complexity and generalization still fall short of humans. 
As we showed, when only 400 examples were provided per rule, both diffusion and autoregressive models didn't learn the rule (fail in panel completion and generation). Whereas human subjects can learn these rules with fewer samples, and without observing all the 40 rules. It is interesting to see what inductive biases or pre-training could be done to close the sample complexity and generalization gap between human and current generative models. 
Given the diverse scaling behavior and learning behavior per rule of two families of models, comparative studies with human subjects could potentially adjudicate between the models and answer which algorithm is behaviorally more similar to humans. 

\paragraph{Advanced reasoning techniques in autoregressive model}
In our study, we focus on autoregressive models in their most basic form. We haven't explored more advanced and intricate designs for autoregressive models. Many recent ideas may be interesting to try to improve their performance, for example, more informative spatial positional encoding \cite{liu2021swin,chu2021twins}; autoregression across spatial scales or Fourier modes instead of across patches \cite{tian2024VAR}. Further, we haven't explored sampling techniques designed for enhancing reasoning for autoregressive models esp. LLMs, e.g. chain-of-thoughts \cite{wei2022CoTReasoning}. We hope our work motivates further investigations into these models and their limitations in reasoning capability. 

\begin{ack}
  Many thanks to Hidenori Tanaka, Martin Wattenberg, Michael Albergo, Andy Keller, Ekdeep Singh Lubana, Talia Konkle, George Alvarez, and Sham Kakade for helpful discussions and insightful feedback. 
  We appreciate the Kempner Institute for the Study of Natural and Artificial Intelligence at Harvard University for providing funding, computing resources, and space for this research; we also thank the Swartz Foundation, ONR grant No. N0014-23-1-2051, and a generous gift from Amazon. 
\end{ack}
\clearpage

\bibliography{iclr2025_conference}
\bibliographystyle{plainnat}


\appendix

\section{Detailed Methods}

\subsection{Dataset construction}\label{method:data_gen_proc}
We have a procedural algorithm to generate rows following the 40 rules. The basic workflow is the following: 
\begin{enumerate}
    \item Choosing the rule-following dimension (Shape, Size, Color, Number, Position);
    \item Choosing the rule type (Constant, Progression, Arithmetic, XOR, OR, AND); 
    \item Switching based on the type of rules
    \begin{itemize}
        \item If it's Constant, Progression, or Arithmetic of object attributes (shape, size, color), then each panel will only have a single value for the attribute, i.e., objects all have the same size in one panel. We will choose three values $(X_1, X_2, X_3)$ that follow the rule (e.g., Progression +2 will be $X_1=1, X_2=3, X_3=5$). 
        \item If it's XOR, OR, AND of object attributes, then each panel will have a set of attributes which suffice the logic operation. Decide the number of objects of each attribute value. 
    \end{itemize}
    \item Choose the other attribute dimensions of the object. 
    \item Check if the other attribute follows other rules, if so, regenerate the other attributes
\end{enumerate}

We synthesized 1,200,000 rows per rule. Three rows following the same rule will be randomly combined into a Raven's Progression Matrix sample (400,000 samples per rule). From this base dataset, we will subsample 400, 4000, 40000, and 400000 samples to construct training sets for the data scaling experiments. By default, we will use the 4000 samples to train most of the models. 
These RPMs are encoded as a $3\times9\times9$ integer matrix. We used $[1.5, 2.5, 2.5]$ and $[2.5, 3.5, 3.5]$ as the mean and std of the three channels to normalize the RPM encoding tensors. 

\subsection{Diffusion model and training details} \label{method:diffusion_training} \label{method:model_spec}

\subsubsection{EDM/UNet}
Our implementation was based on the original EDM code base \citep{karras2022elucidating} and its simplified version \url{https://github.com/yuanzhi-zhu/mini_edm}. 

For model scaling, we tested EDM models with three scales, EDMx1, EDMx2, and EDMx3, which double and triple the width and depth of the UNet model. All UNet models have 3 resolution blocks $(9, 3, 1)$; at each resolution level, the channel numbers are increased by a multiplier $(1,2,4)$. At each resolution, there are self-attention modules. 
Specifically, \texttt{EDMx1},\texttt{EDMx2},\texttt{EDMx3}, used base channel number 64, 128 and 192; while they used 1, 2, and 3 Residual Layer per resolution. 


\subsubsection{DiT}
Our implementation was based on the original DiT code base \citep{peebles2023scalableDiT}, \url{https://github.com/facebookresearch/DiT}

For the major part of the paper, we used patch size 1, which treats each object as a token, with 81 tokens. In the ablation experiments, we also tested the model with patch size 3, which treats each panel as a token, with 9 tokens. The panel tokenization trains much faster but didn't perform as well. 

For model scaling, we adapted the configuration from DiT and SiT and used their Small (-S) (12 layers, 384 hidden dimensions, 6 heads) and Big (-B) scale (12 layers, 768 hidden dimensions, 12 heads). 
Other hyperparameters: 

\subsubsection{SiT} 
Our implementation was based on the original SiT code base \citep{ma2024sit}, \url{https://github.com/willisma/SiT}. 

The network architecture of SiT is almost identical to DiT, with the same object tokenization and configuration for Small and Big scales. 
For the training configuration, we used Velocity prediction, and Linear Path and no loss weighting.

\subsubsection{Default samplers}
By default, during model training, we used deterministic samplers to generate samples for efficiency: Heun's 2nd order sampler with 18 steps for UNet model \citep{karras2022elucidatingEDM}; DDIM sampler with 100 steps for DiT \citep{song2020DDIM}; off-the-shelf Runge-Kutta sampler of order 5 (\texttt{dpori5}) for SiT \citep{ma2024sit}. 

\subsection{Autoregressive models baseline} \label{sec:GPT2_baseline}
For all auto-regressive models, we treated each object as a token. The three integer attributes of each token were embedded through separated embedding layers with 1/3 hidden dimensions and then concatenated as the token embedding. During decoding, we chunked the latent state's output from the final layer into 3 parts and decoded them separately into the three integer attributes of the next token. All the training samples were pre-pended with a \texttt{[SOS]} token with a learned embedding vector. These models are trained on the next token prediction objective. 

\subsubsection{GPT2} 
Our implementation was based on the \texttt{GPT2model} from huggingface \texttt{transformers} library \url{https://github.com/huggingface/transformers/blob/v4.46.0/src/transformers/models/gpt2/modeling_gpt2.py}. We trained GPT2 models at Small, Base, Medium scale. GPT2-S has 384d hidden states, 6 attention heads and 12 layers; GPT2-B has 768d hidden states, 12 attention heads and 12 layers; GPT2-M has 768d hidden states, 12 attention heads and 24 layers.

\subsubsection{Mamba}
Our implementation was based on the original \url{https://github.com/state-spaces/mamba} code base. We used Mamba1 blocks in the backbone. We trained Mamba models at Base, Medium, Big, and Huge scale. Mamba-Base has 768d hidden states and 12 layers; Mamba-Medium has 768d hidden states and 24 layers; Mamba-Big has 1152d hidden states and 36 layers; Mamba-Huge has 1536d hidden states and 48 layers. 

\subsubsection{Training details}
We used AdamW optimizer, with a linear learning rate schedule with 100 warmup steps and a base learning rate 1E-4. Both GPT2 and Mamba models were trained with batch size 64. GPT2s were trained with 1000000 gradient update steps and Mamba by 250000 gradient steps. 

\subsubsection{Sampling details}
For unconditional generation, we sampled the next token with temperature 1.0.

\subsection{Image conditional sampling method (inpainting)}
For diffusion models, we used the Twisted Diffusion Sampler based on their official implementation \url{https://github.com/blt2114/twisted_diffusion_sampler}. 
We compared that to a custom implementation of the Repaint method combined with Heun (for EDM) and DDIM sampler (for DiT). 

For autoregressive models (GPT2, Mamba), we used the greedy sampling with the first 8 panels (and [SOS] token) as their "prompt". 



\newpage

\end{document}